\documentclass[11pt,a4paper]{article}
\usepackage[hyperref]{acl2020}
\usepackage{times}
\usepackage{latexsym}

\usepackage{microtype}

\newcommand{\ACRO}[1]{\textsc{#1}}
\newcommand{\GENIA}{\ACRO{genia}}
\newcommand{\ACEFOUR}{\ACRO{ace 2004}}
\newcommand{\ACEFIVE}{\ACRO{ace 2005}}
\newcommand{\CONLLTWO}{\ACRO{conll 2002}}
\newcommand{\CONLLTHREE}{\ACRO{conll 2003}}
\newcommand{\ONTONOTES}{\ACRO{ontonotes}}

\usepackage{amsmath}
\usepackage{booktabs}
\usepackage{footnote}

\DeclareMathOperator*{\argmax}{arg\,max}
\usepackage{multirow}
\usepackage{graphicx}

\aclfinalcopy 

\title{Named Entity Recognition as Dependency Parsing}

\author{Juntao Yu \\
  Queen Mary University \\
  London, UK \\
  \texttt{juntao.yu@qmul.ac.uk} \\\And
  Bernd Bohnet \\
  Google Research \\
  Netherlands \\
  \texttt{bohnetbd@google.com} \\\And
  Massimo Poesio\\
  Queen Mary University \\
  London, UK \\
  \texttt{m.poesio@qmul.ac.uk} \\}

\date{}

\begin{document}
\maketitle
\begin{abstract}

Named Entity Recognition (NER) is a fundamental task in Natural Language Processing, concerned with identifying spans of text expressing references to entities. NER research  is often focused on  flat entities only (flat NER), ignoring the fact that entity references can be nested, as in \textit{[Bank of [China]]} \cite{finkel-manning-2009-nested}. 
In this paper, we use ideas from graph-based dependency parsing to provide our model a global view on the input via a biaffine model \cite{dozat-and-manning2017-parser}. 
The biaffine model scores pairs of start and end tokens in a sentence which we use to explore all spans, so that the model is able to predict named entities accurately. 
We show 
that the model works well for both nested and flat NER 
through evaluation 
on 8 corpora and achieving SoTA performance on all of them, with accuracy gains of up to 2.2 percentage points. 

\end{abstract}

\section{Introduction}

`Nested Entities' are named entities containing references to other named entities as in 
\textit{[Bank of [China]]}, in which both \textit{[China]} and \textit{[Bank of China]} are named entities.
Such nested entities are frequent in data sets like {\ACEFOUR}, {\ACEFIVE} and {\GENIA} (e.g., 17\% of NEs in {\GENIA} are nested \cite{finkel-manning-2009-nested}, 
altough  the more widely used set such as {\CONLLTWO}, 2003 and {\ONTONOTES} only contain  so called flat named entities and nested entities are ignored. 

The current SoTA models all adopt 
a neural network architecture without hand-crafted features,
which makes them more adaptable to different tasks, languages and domains 
\cite{glample2016-ner,chiu2016named,peters2018elmo,devlin2019bert,ju-etal-2018-neural,sohrab-miwa-2018-deep,strakova-etal-2019-neural}. 
In this paper, we introduce a method to handle both types of NEs in one system by adopting ideas from the biaffine dependency parsing model of \newcite{dozat-and-manning2017-parser}. 
For dependency parsing, the system predicts a head for each token and assigns a relation to the head-child pairs. 
In this work, we reformulate NER as the task of identifying start and end indices, as well as assigning a category to the span defined by these pairs. 
Our system uses a biaffine model on top of a multi-layer BiLSTM to assign scores to all possible spans in a sentence. After that, instead of building dependency trees, we rank the candidate spans by their scores and return the top-ranked spans that comply with constraints for flat or nested NER. 
We evaluated our system on three nested NER benchmarks ({\ACEFOUR}, {\ACEFIVE}, {\GENIA}) and five flat NER corpora ({\CONLLTWO} (Dutch, Spanish) {\CONLLTHREE} (English, German), and {\ONTONOTES}). 
The results show that our system achieved SoTA results on all three nested NER corpora, and on all five flat NER corpora with substantial gains of up to 2.2\% absolute percentage points compared to the previous SoTA. We provide the code as open source\footnote{The code is available at \url{https://github.com/juntaoy/biaffine-ner}}.

\section{Related Work}

\textbf{Flat Named Entity Recognition.} 
The majority of flat NER models are based on a sequence labelling approach.
\newcite{collobert2011natural} introduced a neural NER model that uses CNNs to encode tokens combined with a CRF layer for the classification. 
Many other neural systems followed this approach but used instead LSTMs to encode the input and a CRF for the prediction \cite{glample2016-ner,ma-hovy-2016-end,chiu2016named}. 
These latter models were later extended to use context-dependent embeddings such as ELMo \cite{peters2018elmo}. 
\newcite{clark-etal-2018-semi}  quite successfully used cross-view training (CVT) paired with multi-task learning. This method yields impressive gains for a number of NLP applications including NER.
\newcite{devlin2019bert} invented BERT, a  bidirectional transformer architecture for the training of language models. BERT and its siblings provided better language models that turned again into higher scores for NER.
 
\newcite{glample2016-ner} cast NER as transition-based dependency parsing using a Stack-LSTM. They compare with a LSTM-CRF model which turns out to be a very strong baseline. Their transition-based system uses two transitions (shift and reduce) to mark the named entities and handles flat NER while our system has been designed to handle both nested and flat entities. 

\begin{figure}[t]
    \centering
    \includegraphics[width=.85\columnwidth]{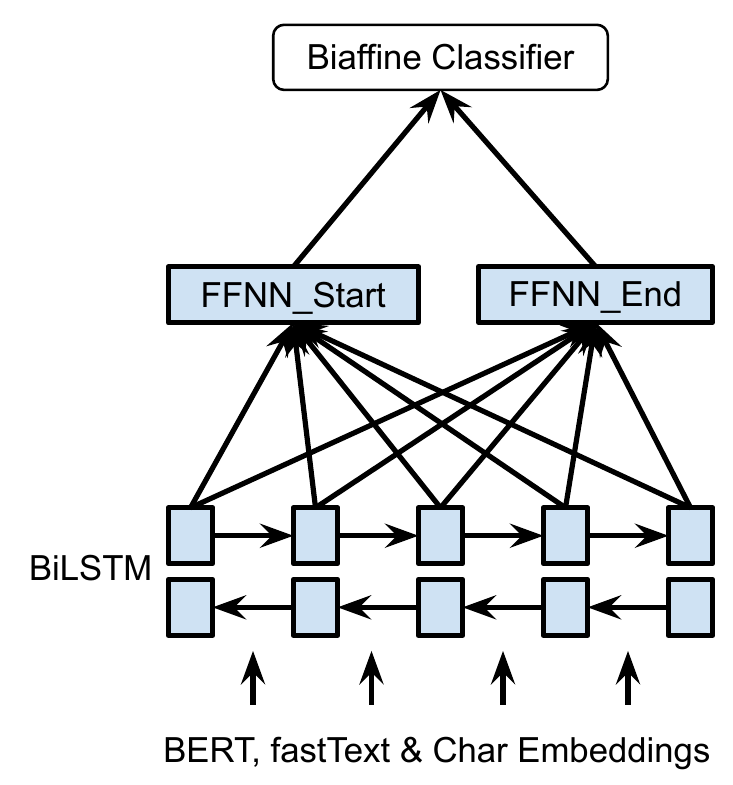}
    \caption{The network architectures of our system.}
    \label{fig:nngraph}
\end{figure}

\textbf{Nested Named Entity Recognition}. 
Early work on nested NER, motivated particularly by the GENIA corpus, includes \cite{shen-et-al:BIONLP03,alex-et-al:BIONLP07,finkel-manning-2009-nested}. 
\newcite{finkel-manning-2009-nested} also proposed a constituency parsing-based approach.
In the last years, we saw an increasing number of neural models targeting nested NER as well. 
\newcite{ju-etal-2018-neural} suggested a LSTM-CRF model to predict nested named entities. Their algorithm iteratively continues until no further entities are predicted. \newcite{lin-etal-2019-sequence} tackle the problem in two steps: they first detect the entity head, and then they infer the entity boundaries as well as the category of the named entity. 
\newcite{strakova-etal-2019-neural} tag the nested named entity by a sequence-to-sequence model exploring combinations of context-based embeddings such as ELMo, BERT, and Flair.  \newcite{zheng2019boundary} use a boundary aware network to solve the nested NER. Similar to our work, \newcite{sohrab-miwa-2018-deep} enumerate exhaustively all possible spans up to a defined length by concatenating the LSTMs outputs for the start and end position and then using this to calculate a score for each span. Apart from the different network and word embedding configurations, the main difference between their model and ours is there for the use of biaffine model. Due to the biaffine model, we get a global view of the sentence while \newcite{sohrab-miwa-2018-deep} concatenates the output of the LSTMs of possible start and end positions up to a distinct length. \newcite{dozat-and-manning2017-parser} demonstrated that the biaffine mapping performs significantly better than just the concatenation of pairs of LSTM outputs.

\section{Methods}
Our model is inspired by the dependency parsing model of \newcite{dozat-and-manning2017-parser}. 
We use both word embeddings and character embeddings as input, 
and feed the output into a BiLSTM and finally to a biaffine classifier. 

Figure \ref{fig:nngraph} shows an overview of the architecture. To encode words, we use both BERT$_{Large}$ and fastText embeddings \cite{bojanowski2016enriching}. For BERT we follow the recipe of \cite{kantor2019bertee} to obtain the context dependent embeddings for a target token with 64 surrounding tokens each side. 
For the character-based word embeddings, we use a CNN to encode the characters of the tokens. The concatenation of the word and character-based word embeddings is feed into a BiLSTM to obtain the word representations ($x$).

After obtaining the word representations from the BiLSTM, we apply two separate FFNNs to create different representations ($h_s/h_e$) for the start/end of the spans. Using different representations for the start/end of the spans allow the system to learn 
to identify the start/end of the spans separately. This improves accuracy compared to the model which directly uses the outputs of the LSTM since the context of the start and end of the entity are different. Finally, we employ a biaffine model over the sentence to create a $l \times l \times c$ scoring tensor ($r_m$), where $l$ is the length of the sentence and $c$ is the number of NER categories $+\ 1$(for non-entity). 
We compute the score for a span $i$ by:

\vspace{-10pt}
\begin{align*}
h_{s}(i) =\ & \textsc{FFNN}_{s}(x_{s_i})\\
h_{e}(i) =\ & \textsc{FFNN}_{e}(x_{e_i})\\
r_m(i) =\ & h_{s}(i)^\top \text{U}_{m} h_{e}(i)\\
&+ W_m (h_{s}(i) \oplus h_{e}(i)) + b_{m}
\end{align*}
where $s_i$ and $e_i$ are the start and end indices of the span $i$, $\text{U}_{m}$ is a $d \times c \times d$ tensor, $W_m$ is a $2d \times c$ matrix and $b_{m}$ is the bias. 

The tensor $r_m$ provides scores for all possible spans that could constitute a named entity under the constrain that $s_i \leq e_i$ (the start of entity is before its end).
We assign each span a NER category $y'$:

\vspace{-5pt}
$$y'(i) = \argmax~r_m(i)$$
We then rank all the spans that have a category other than "non-entity" by their category scores ($r_m(i_{y'})$) in descending order and apply following post-processing constraints: For nested NER, a entity is selected as long as it does not {\it clash} the boundaries of higher ranked entities. We denote a entity $i$ to {\it clash} boundaries with another entity $j$ if $s_i < s_j \leq e_i < e_j$ or $s_j < s_i \leq e_j < e_i$, e.g. in \textit{the Bank of China}, the entity \textit{the Bank of} clashes boundary with the entity \textit{Bank of China}, hence only the span with the higher category score will be selected. For flat NER, we apply one more constraint, in which any entity containing or is inside an entity ranked before it will not be selected. 
The learning objective of our named entity recognizer is to assign a correct category (including the non-entity) to each valid span. Hence it is a multi-class classification problem and we optimise our models with softmax cross-entropy:

$$p_m(i_c) = \frac{exp(r_m(i_c))}{\sum^{C}_{\hat{c}=1} exp(r_m(i_{\hat{c}}))}$$
$$loss = - \sum_{i=1}^N\sum_{c=1}^C y_{i_c} \log p_m(i_c)$$



\begin{table}[t]
    \centering
    \begin{tabular}{l l}
    \toprule
    \bf Parameter & \bf Value \\
    \midrule
    BiLSTM size & 200\\
    BiLSTM layer &3\\
    BiLSTM dropout&0.4\\
    FFNN size&150\\
    FFNN dropout&0.2\\
    BERT size & 1024\\
    BERT layer&last 4\\
    fastText embedding size&300\\
    Char CNN size&50\\
    Char CNN filter widths&[3,4,5]\\
    Char embedding size&8\\
    Embeddings dropout &0.5\\
    Optimiser & Adam\\
    learning rate& 1e-3\\
    \bottomrule
    \end{tabular}
    \caption{Major hyperparameters for our models.}
    \label{tab:hyperparameters}
\end{table}

\section{Experiments}
\textbf{Data Set}. We evaluate our system on both nested and flat NER, for the nested NER task, we use the {\ACEFOUR}\footnote{https://catalog.ldc.upenn.edu/LDC2005T09}, {\ACEFIVE}\footnote{https://catalog.ldc.upenn.edu/LDC2006T06}, and  {\GENIA} \cite{kim2003genia} corpora; for flat NER, we test our system on the {\CONLLTWO} \cite{tjong-kim-sang-2002-introduction},
{\CONLLTHREE} \cite{tjong-kim-sang-de-meulder-conll2003-introduction} and  {\ONTONOTES}\footnote{https://catalog.ldc.upenn.edu/LDC2013T19} corpora.

For {\ACEFOUR}, {\ACEFIVE} we follow the same settings of \newcite{lu-roth-2015-joint} and \newcite{muis-lu-2017-labeling} to split the data into 80\%,10\%,10\% for train, development and test set respectively. To make a fair comparson we also used the same documents as in \newcite{lu-roth-2015-joint} for each split.

For {\GENIA}, we use the {\GENIA} v3.0.2 corpus. We preprocess the dataset following the same settings of \newcite{finkel-manning-2009-nested} and \newcite{lu-roth-2015-joint} and use 90\%/10\% train/test split. For this evaluation, since we do not have a development set, we train our system on 50 epochs and evaluate on the final model.

For {\CONLLTWO} and {\CONLLTHREE}, we evaluate on all four languages (English, German, Dutch and Spanish). We follow \newcite{glample2016-ner} to train our system on the concatenation of the train and development set. 

For {\ONTONOTES}, we evaluate on the English corpus and follow \newcite{strubell-etal-2017-fast} to use the same train, development and test split as used in CoNLL 2012 shared task for coreference resolution \cite{pradhan2012conllst}.

\textbf{Evaluation Metric}. We report recall, precision and F1 scores for all evaluations. The named entity is considered correct when both boundary and category are predicted correctly.

\textbf{Hyperparameters} We use a unified setting for all of the experiments, Table \ref{tab:hyperparameters} shows hyperparameters for our system.

\begin{savenotes}
\begin{table}[t]
\setlength{\tabcolsep}{3pt}
    \centering
    \begin{tabular}{llll}
    \toprule
    \bf Model & \bf P &\bf R&\bf F1 \\
    \midrule
    \multicolumn{4}{c}{\ACEFOUR}\\\midrule
    \newcite{katiyar-cardie-2018-nested}&73.6&71.8&72.7\\
    \newcite{wang-etal-2018-neural-transition}&-&-&73.3\\
    \newcite{wang-lu-2018-neural}&78.0&72.4&75.1\\
    \newcite{strakova-etal-2019-neural}&-&-&84.4\\
    \newcite{luan-etal-2019-general}&-&-&84.7\\
    Our model& 87.3& 86.0&\bf 86.7\\
    \midrule
    \multicolumn{4}{c}{\ACEFIVE}\\\midrule
    \newcite{katiyar-cardie-2018-nested}&70.6&70.4&70.5\\
    \newcite{wang-etal-2018-neural-transition}&-&-&73.0\\
    \newcite{wang-lu-2018-neural}&76.8&72.3&74.5\\
    \newcite{lin-etal-2019-sequence}&76.2&73.6&74.9\\
    \newcite{fisher-vlachos-2019-merge}&82.7&82.1&82.4\\
    \newcite{luan-etal-2019-general}&-&-&82.9\\
    \newcite{strakova-etal-2019-neural}&-&-&84.3\\
    Our model& 85.2& 85.6&\bf 85.4\\
    \midrule
    \multicolumn{4}{c}{\GENIA}\\\midrule
    \newcite{katiyar-cardie-2018-nested}&79.8&68.2&73.6\\
    \newcite{wang-etal-2018-neural-transition}&-&-&73.9\\
    \newcite{ju-etal-2018-neural}&78.5&71.3&74.7\\
    \newcite{wang-lu-2018-neural}&77.0&73.3&75.1\\
    \newcite{sohrab-miwa-2018-deep}\footnote{In \newcite{sohrab-miwa-2018-deep}, the last 10\% of the training set is used as a development set, we include their result mainly because their system is similar to ours.}&93.2&64.0 &77.1\\
    \newcite{lin-etal-2019-sequence}&75.8&73.9&74.8\\
    \newcite{luan-etal-2019-general}&-&-&76.2\\
    \newcite{strakova-etal-2019-neural}&-&-&78.3\\
    Our model& 81.8& 79.3&\bf 80.5\\
    \bottomrule
    \end{tabular}
    \caption{State of the art comparison on {\ACEFOUR}, {\ACEFIVE} and {\GENIA} corpora for nested NER.}
    \label{tab:nested_ner}
\end{table}
\end{savenotes}

\begin{savenotes}
\begin{table}[th]
\renewcommand{\arraystretch}{0.9}
\setlength{\tabcolsep}{3pt}
    \centering
    \begin{tabular}{llll}
    \toprule
    \bf Model & \bf P &\bf R&\bf F1 \\
    \midrule
    \multicolumn{4}{c}{\ONTONOTES}\\\midrule
    \newcite{chiu2016named}&86.0&86.5&86.3\\
    \newcite{strubell-etal-2017-fast}&-&-&86.8\\
    \newcite{clark-etal-2018-semi}&-&-&88.8\\
    \newcite{fisher-vlachos-2019-merge}&-&-&89.2\\
    Our model&91.1&91.5&\bf 91.3\\\midrule
    \multicolumn{4}{c}{{\CONLLTHREE} English}\\\midrule
    \newcite{chiu2016named}&91.4&91.9&91.6\\
    \newcite{glample2016-ner}&-&-&90.9\\
    \newcite{strubell-etal-2017-fast}&-&-&90.7\\
    \newcite{devlin2019bert}&-&-& 92.8\\
    \newcite{strakova-etal-2019-neural}&-&-& 93.4\\
    Our model&93.7&93.3&\bf 93.5\\
    \midrule
    \multicolumn{4}{c}{{\CONLLTHREE} German}\\\midrule
    \newcite{glample2016-ner}&-&-&78.8\\
    \newcite{strakova-etal-2019-neural}&-&-&85.1\\
    Our model&88.3&84.6&\bf 86.4\\\midrule
    \multicolumn{4}{c}{{\CONLLTHREE} German revised\footnote{The revised version is provided by the shared task organiser in 2006 with more consistent annotations. We confirmed with the author of \newcite{akbik-etal-2018-flair} that they used the revised version.}}\\\midrule
        \newcite{akbik-etal-2018-flair}&-&-& 88.3\\
        Our model & 92.4& 88.2& \bf 90.3\\
    \midrule
    \multicolumn{4}{c}{{\CONLLTWO} Spanish}\\\midrule
    \newcite{glample2016-ner}&-&-&85.8\\
    \newcite{strakova-etal-2019-neural}&-&-&88.8\\
    Our model &90.6&90.0&\bf 90.3\\
    \midrule
    \multicolumn{4}{c}{{\CONLLTWO} Dutch}\\\midrule
    \newcite{glample2016-ner}&-&-&81.7\\
    \newcite{akbik-etal-2019-pooled}&-&-&90.4\\
    \newcite{strakova-etal-2019-neural}&-&-& 92.7\\
    Our model &94.5&92.8&\bf 93.7\\
    \bottomrule
    \end{tabular}
    \caption{State of the art comparison on {\CONLLTWO}, {\CONLLTHREE}, {\ONTONOTES} corpora for flat NER.}
    \label{tab:flat_ner}
\end{table}
\end{savenotes}

\section{Results on Nested NER}
Using the constraints for nested NER, we first evaluate our system on nested named entity corpora: {\ACEFOUR}, {\ACEFIVE} and {\GENIA}. Table \ref{tab:nested_ner} shows the results. Both {\ACEFOUR} and {\ACEFIVE} contain 7 NER categories and have a relatively high ratio of nested entities (about 1/3  of then named entities are nested). Our results outperform the previous SoTA system by 2\% ({\ACEFOUR}) and 1.1\% ({\ACEFIVE}), respectively. 
{\GENIA} differs from {\ACEFOUR} and {\ACEFIVE} and uses five medical categories such as DNA or RNA. For the {\GENIA} corpus our system achieved an F1 score of 80.5\% and improved the SoTA by 2.2\% absolute. Our hypothesise is that for {\GENIA} the high accuracy gain is due to our structural prediction approach and that sequence-to-sequence models rely more on the language model embeddings which are less informative for categories such as DNA, RNA. Our system achieved SoTA results on all three corpora for nested NER and demonstrates well the advantages of a structural prediction over sequence labelling approach.

\section{Results on Flat NER}
We evaluate our system on five corpora for flat NER ({\CONLLTWO} (Dutch, Spanish), {\CONLLTHREE} (English, German) and {\ONTONOTES}. Unlike most of the systems that treat flat NER as a sequence labelling task, our system predicts named entities by considering all possible spans and ranking them. The {\ONTONOTES} corpus consists of documents form 7 different domains and is annotated with 18 fine-grained named entity categories. To predict named entities for this corpus is more difficult than for {\CONLLTWO} and {\CONLLTHREE}. These corpora use coarse-grained named entity categories (only 4 categories). The sequence-to-sequence models usually perform better on the {\CONLLTHREE} English corpus (see Table \ref{tab:flat_ner}), e.g. the system of \newcite{chiu2016named,strubell-etal-2017-fast}. In contrast, our system is less sensitive to the domain and the granularity of the categories. As shown in Table \ref{tab:flat_ner}, our system achieved an F1 score of 91.3\% on the {\ONTONOTES} corpus and is very close to our system performance on the {\CONLLTHREE} corpus (93.5\%). 
On the multi-lingual data, our system achieved F1 scores of 86.4\% for German, 90.3\% for Spanish and 93.5\% for Dutch. Our system outperforms the previous SoTA results by large margin of 2.1\%, 1.5\%, 1.3\% and 1\% on {\ONTONOTES}, Spanish, German and Dutch corpora respectively and is slightly better than the SoTA on English data set. In addition, we also tested our system on the revised version of German data to compare with the model by \newcite{akbik-etal-2018-flair}, our system again achieved a substantial gain of 2\% when compared with their system.

\begin{table}[t]
\centering
\begin{tabular}{l l l}
\toprule
&\bf F1& \bf $\Delta$\\
\midrule
Our model&89.9&\\
\ \ - biaffine&89.1&0.8\\
\ \ - BERT emb&87.5&2.4\\
\ \ - fastText emb&89.5&0.4\\
\ \ - Char emb&89.8&0.1\\
\bottomrule
\end{tabular}
\caption{\label{table:analysis} The comparison between our full model and ablated models on {\ONTONOTES} development set.}
\end{table}

\section{Ablation Study}
To evaluate the contribution of individual components of our system, we further remove selected components and use {\ONTONOTES} for evaluation (see Table \ref{table:analysis}). We choose {\ONTONOTES} for our ablation study as it is the largest corpus.  

\textbf{Biaffine Classifier} We  replace the biaffine mapping with a CRF layer and convert our system into a sequence labelling model. The CRF layer is frequently used in models for flat NER, e.g. \cite{glample2016-ner}. When we replace the biaffine model of  our system with a CRF layer, the performance drops by 0.8 percentage points (Table \ref{table:analysis}). The large performance difference shows the benefit of adding a biaffine model and confirms our hypothesis that the dependency parsing framework is an important factor for the high accuracy of our system. 

\textbf{Contextual Embeddings} We ablate BERT embeddings and as expected, after removing BERT embeddings, the system performance drops by a large number of 2.4 percentage points (see Table \ref{table:analysis}). This shows that BERT embeddings are one of the most important factors for the accuracy.

\textbf{Context Independent Embeddings} We remove the context-independent fastText embedding from our system. The context-independent embedding contributes 0.4\% towards the score of our full system (Table \ref{table:analysis}). Which suggests that even with the BERT embeddings enabled, the context-independent embeddings can still make quite noticeable improvement to a system. 

\textbf{Character Embeddings} Finally, we remove the character embeddings. As we can see from Table \ref{table:analysis}, the impact of character embeddings is quite small. One explanation would be that English is not a morphologically rich language hence does not benefit largely from character-level information and the BERT embeddings itself are based on word pieces that already capture some character-level information.

Overall, the biaffine mapping and the BERT embedding together contributed most to the high accuracy of our system.

\section{Conclusion}
In this paper, we reformulate NER as a structured prediction task and adopted a SoTA dependency parsing approach for nested and flat NER. Our system uses contextual embeddings as input to a multi-layer BiLSTM. We employ a biaffine model to assign scores for all spans in a sentence. Further constraints are used to predict nested or flat named entities. We evaluated our system on eight named entity corpora. The results show that our system achieves SoTA on all of the eight corpora. We demonstrate that advanced structured prediction techniques lead to substantial improvements for both nested and flat NER.

\section*{Acknowledgments}
This research was supported in part by the DALI project, ERC Grant 695662.

\newpage
\bibliography{acl2020}
\bibliographystyle{acl_natbib}

\end{document}